\documentclass[lettersize,journal]{IEEEtran}
\usepackage{amsmath,amsfonts,mathtools}
\usepackage{algorithmic}
\usepackage{algorithm}
\usepackage{array}
\usepackage[caption=false,font=footnotesize,textfont=rm]{subfig}
\usepackage{textcomp}
\usepackage{stfloats}
\usepackage{url}
\usepackage{verbatim}
\usepackage{graphicx}
\usepackage{cite}
\usepackage{bm}
\usepackage{float}
\usepackage{xcolor}

\newcommand{\vect}[1]{\bm{#1}}
\newcommand{\mat}[1]{\bm{#1}}
\newcommand{\R}{\mathbb{R}}
\newcommand{\T}{^{\top}}
\newcommand{\argmin}{\operatorname*{arg\,min}}

\begin{document}

\title{Force-Aware Reinforcement Learning with Hybrid Sensorless Force Estimation for Wheeled-Legged Loco-Manipulation}

\author{Xuanqi Zeng, Jiaming Wang, Tianlin Zhang, Lingwei Zhang, Botian Xu, Weipeng Xia, \\ Zhongyu Li, Yun-Hui Liu,~\IEEEmembership{Fellow,~IEEE}
\thanks{ This work is supported in part by the CUHK T Stone Robotics Institute and in part by the InnoHK initiative of the Innovation and Technology Commission of the Hong Kong Special Administrative Region Government via the Hong Kong Centre for Logistics Robotics. (\textit{Corresponding author: Yun-Hui Liu.})

All the authors are with the Department of Mechanical and Automation Engineering at the Chinese University of Hong Kong, Shatin, Hong Kong. (e-mail: yhliu@cuhk.edu.hk).}
\thanks{}}




\maketitle

\begin{abstract}
Force-controlled loco-manipulation requires a whole-body policy to coordinate locomotion and arm motion while regulating end-effector interaction forces. This is challenging under floating-base dynamics and changing support contacts, particularly when end-effector force/torque sensing is unavailable for control. This paper presents a force-aware reinforcement learning approach with hybrid sensorless force estimation for wheeled-legged loco-manipulation. The proposed method provides a structured estimate of the end-effector force as an explicit policy observation, enabling force-guided contact behavior without using an end-effector force/torque sensor for control. The force estimate is obtained by combining generalized momentum observation, contact-constrained wrench projection, and temporal residual learning: the model-based components extract the physically structured part of the whole-body disturbance, while the residual network compensates the remaining motion-dependent bias. The estimated force is integrated into a mode-conditioned whole-body policy with an axis-wise force/position selector, allowing free-space motion, pure force regulation, and hybrid force/position control within one controller. Simulation results demonstrate improved sensorless force estimation and force-control performance. Hardware experiments further validate the proposed controller through quantitative valve-rotation and hybrid wiping evaluations, together with force-guided door opening and zero-force human-guided motion on a real wheeled-legged platform.
\end{abstract}


\section{Introduction}

\begin{figure}[t]
\centering
\includegraphics[width=\columnwidth]{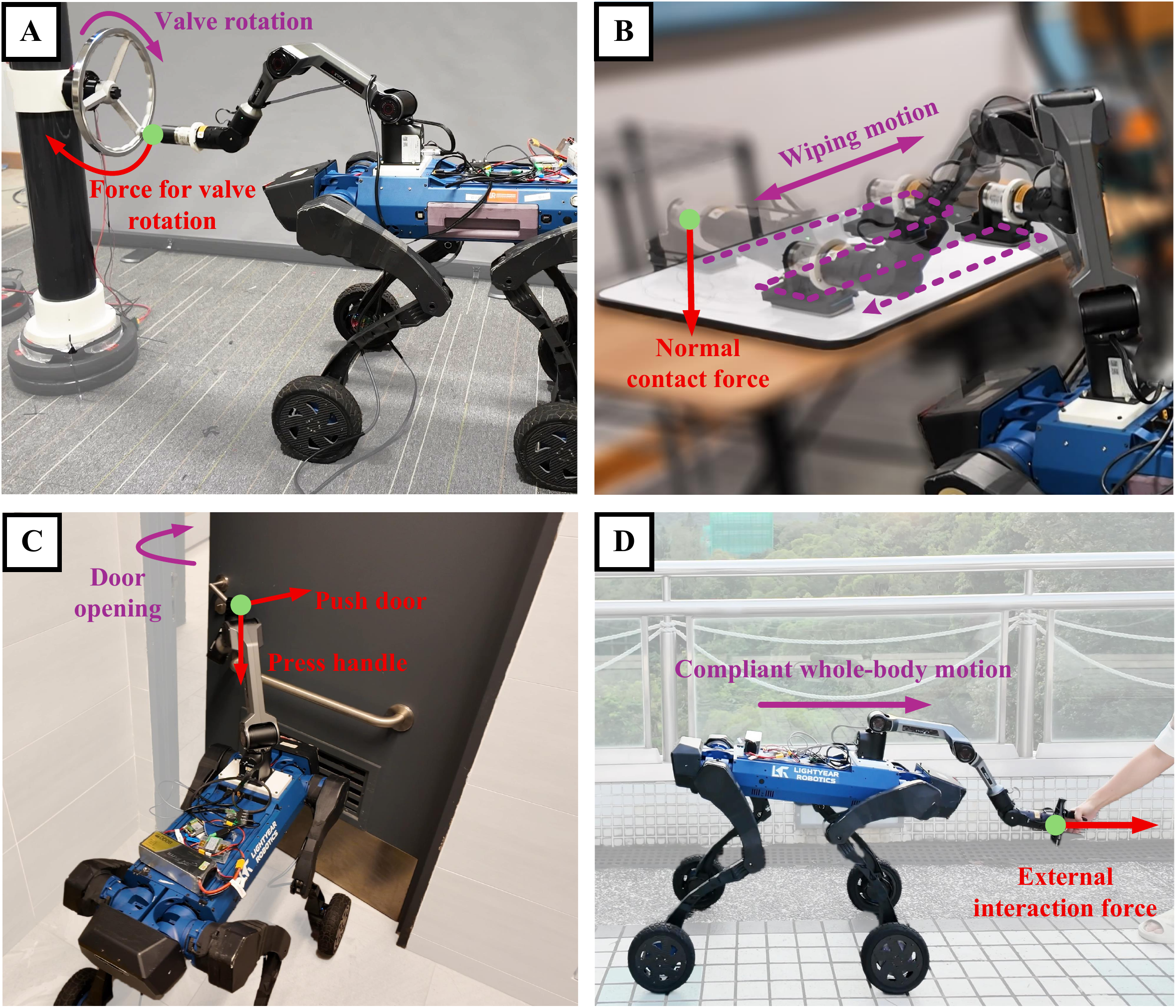}
\caption{Hardware demonstrations of the proposed force-aware controller:
(A) valve rotation with whole-body motion,
(B) hybrid force/position board wiping,
(C) opening a spring-loaded door by force control, and
(D) compliant zero-force/human-guidance interaction. These tasks require the robot to coordinate locomotion, arm motion, and physical interaction with the environment.}
\label{fig:task_overview}
\vspace{-0.3cm}
\end{figure}


RL-based force-control loco-manipulation aims to enable legged mobile manipulators to change the state of objects and environments through learned whole-body contact interaction. The hardware demonstrations previewed in Fig.~\ref{fig:task_overview} illustrate the range of force-controlled loco-manipulation behaviors considered in this work, including valve rotation, hybrid force/position wiping, force-guided door opening, and zero-force human-guided motion. 
Recent learning-based loco-manipulation methods have made substantial progress in coordinating locomotion and arm motion for end-effector position or pose tracking~\cite{ma2022combining,fu2023deep,ferrolho2023roloma,liu2025visual,portela2025wholebody,jiang2024learning,hou2025efficient}. 
However, contact-rich tasks require more than pose tracking: classical force, impedance, and hybrid force/position control studies have long shown that manipulation under environmental constraints requires regulating physical interaction in addition to motion~\cite{mason1981compliance,raibert1981hybrid,hogan1985impedance}. 
For legged loco-manipulation, this requirement becomes more difficult because the robot must generate task-effective contact while simultaneously maintaining floating-base stability and support contacts~\cite{rigo2023contact,portela2024force,zhi2025unified}. 
Thus, the key problem studied in this paper is how an RL-based wheeled-legged manipulator can obtain and use a physically meaningful representation of end-effector interaction for force-guided loco-manipulation.

\begin{figure}[htp]
\centering
\includegraphics[width=3.5in]{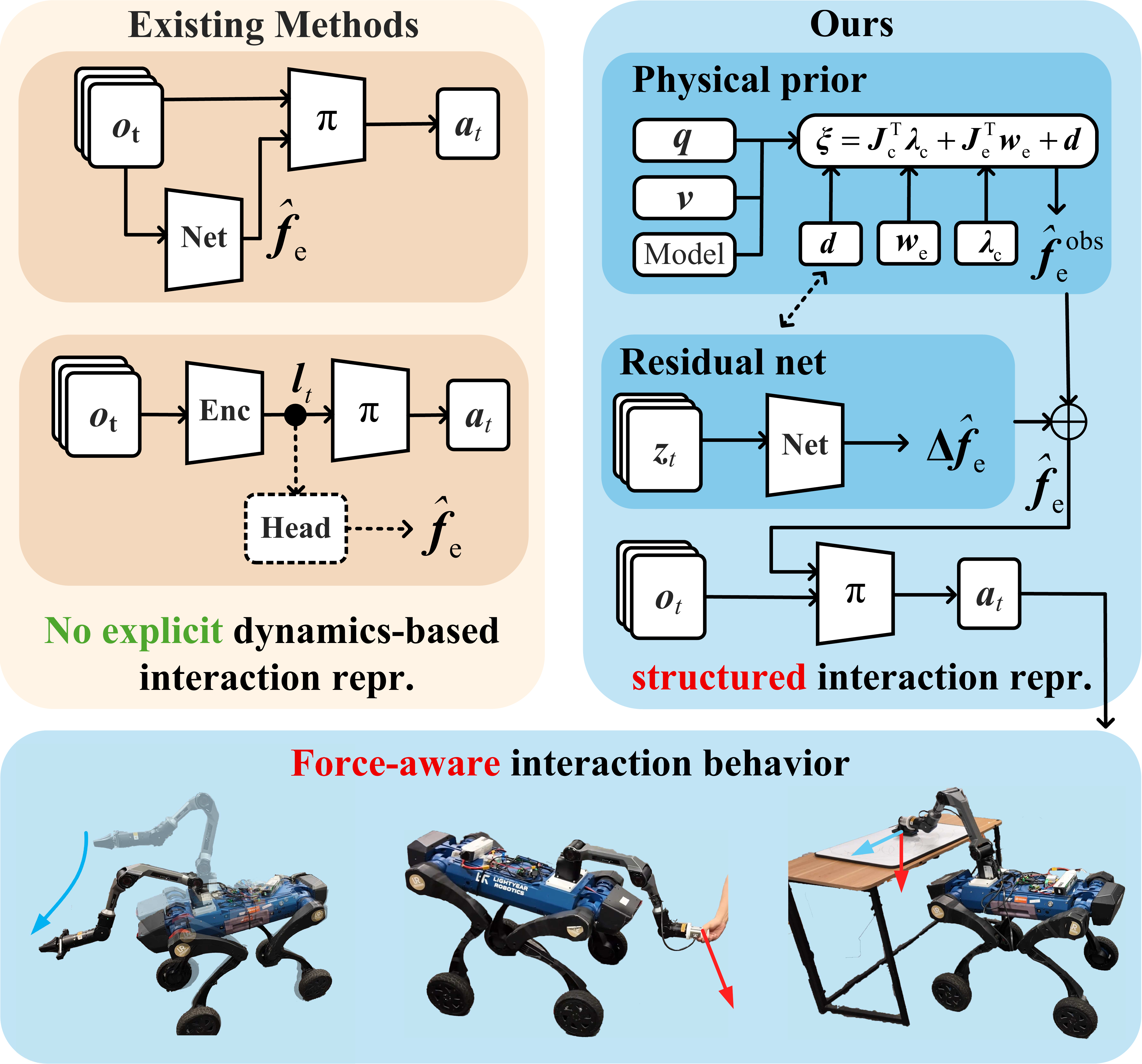}
\caption{
Conceptual comparison and simplified pipeline for force-controlled loco-manipulation.
Existing methods represent interaction through learned force
estimates~\cite{portela2024force} or history features with or without
auxiliary force prediction~\cite{zhi2025unified,facet2025force}.
The proposed method combines a physics-prior estimate
$\hat{\vect{f}}_{\mathrm{e}}^{\mathrm{obs}}$
with a learned residual
$\Delta\hat{\vect{f}}_{\mathrm{e}}$
to obtain a structured force estimate
$\hat{\vect{f}}_{\mathrm{e}}$.
The force estimation supports more task-effective position, force, and hybrid force/position behaviors.
}
\vspace{-0.3cm}
\label{fig:interaction_representation}
\end{figure}

Recent works have begun to study learned force control and unified force/position control for legged manipulators~\cite{portela2024force,zhi2025unified,facet2025force}. These works show that reinforcement learning can generate useful force-related behaviors, but they also reveal that force-control loco-manipulation is still limited by how the interaction state is represented.  One direction learns whole-body force-control policies through direct force-tracking objectives, with learned gripper-force estimates supplied to the actor alongside robot observations and commands~\cite{portela2024force}. This enables the robot to execute force commands through coordinated whole-body actions, but accurate regulation still depends on the learned feedback and policy handling the coupled effects of base motion, support-contact variation, actuator response, and environment reaction.
Another direction unifies force and position control through impedance-derived motion-tracking objectives, with shared history features supplied to the actor and a state-estimation head~\cite{zhi2025unified}. Although this enables unified force/position behavior, accurate force regulation depends on the effectiveness of the motion compensation under the encountered contact and sliding dynamics. Similarly, force-adaptive or impedance-reference learning methods improve compliance under external forces, but their motivation also highlights that conventional position/velocity tracking objectives are agnostic to the forces experienced by the robot during forceful interaction~\cite{facet2025force}.
Therefore, existing learning-based force-control methods are promising, but coordinating accurate force regulation with whole-body motion still relies heavily on learned interaction representations, as summarized in Fig.~\ref{fig:interaction_representation}.

Constructing such a force estimate is particularly challenging for floating-base wheeled-legged loco-manipulation.
For fixed-base manipulators, model-based collision detection, external torque observation, and sensorless force estimation have been extensively studied~\cite{deluca2006collision,haddadin2017robot,garofalo2019sliding}. 
Momentum-observer-based methods are attractive because they exploit robot dynamics and proprioceptive sensing to estimate external disturbance without requiring an end-effector force/torque sensor~\cite{deluca2006collision,haddadin2017robot,garofalo2019sliding}. Recent work on legged robots has also shown that contact estimation and disturbance interpretation are themselves nontrivial in floating-base systems~\cite{fink2020proprioceptive,payne2024multimomentum}. 
As a result, neither proprioceptive history nor a raw model-based disturbance residual directly provides the interaction representation needed by the policy. 
The first problem addressed in this paper is therefore how to extract a physically meaningful end-effector interaction estimate from a mixed floating-base whole-body disturbance.

Purely learning-based force or contact estimators provide another possible solution, since temporal models can infer hidden contact-related quantities from proprioceptive history~\cite{osburg2022using,lim2023proprioceptive,shan2024fine,Shan2024_icra}. 
However, such estimators must learn the complete mapping from robot history to interaction force, including rigid-body dynamics, support-contact transitions, actuator effects, friction, and environment response. 
This can make the estimator dependent on broad force-labeled data coverage and simulation fidelity. 
We therefore adopt a hybrid view: model-based estimation should explain the physically structured part of the floating-base interaction, while learning should compensate only the remaining motion-dependent bias.

To address the first problem, we propose a hybrid sensorless force estimator. A generalized momentum observer estimates the whole-body disturbance residual from proprioceptive measurements and commanded joint torques. A contact-constrained wrench projection then uses robot kinematics and support-contact feasibility to jointly estimate the support reactions and end-effector wrench, producing a physically structured nominal interaction estimate. Because model-based decomposition cannot remove all errors caused by inertial mismatch, rolling-contact effects, actuator mismatch, friction, sensing error, and observer transients, a temporal residual model compensates the remaining force-estimation bias.

The second problem is how to use this estimated interaction information in reinforcement learning. 
We provide the corrected end-effector force estimate to a mode-conditioned whole-body policy as an explicit interaction observation. 
An axis-wise Cartesian task mask specifies whether each translational direction is assigned to position tracking or force-guided interaction, allowing one policy to represent free-space motion, pure force interaction, and hybrid force/position loco-manipulation. 
Thus, the proposed method is not intended merely to improve force-estimation accuracy as a standalone objective; it constructs a physically grounded interaction representation and uses it to support learned whole-body contact behavior. The resulting controller is validated on a real wheeled-legged platform in the hardware demonstrations previewed in Fig.~\ref{fig:task_overview}.

The main contributions of this work are summarized as follows:
\begin{enumerate}
\item We propose a force-aware whole-body reinforcement learning
method for wheeled-legged loco-manipulation that incorporates
structured sensorless force observations and axis-wise
force/position tracking objectives into policy learning.
The resulting mode-conditioned policy coordinates locomotion
and arm motion to perform free-space tracking, force regulation,
and hybrid force/position interaction within a single controller.

\item We propose a hybrid sensorless force estimator that combines generalized momentum observation, contact-constrained wrench projection, and temporal residual learning to extract a physically structured end-effector interaction-force estimate from a mixed whole-body disturbance residual.

\item We evaluate the proposed method through force-control
and estimator comparisons in simulation, and validate
sensorless force-controlled loco-manipulation on a real
wheeled-legged platform through valve rotation, hybrid wiping,
force-guided door opening, and zero-force human-guided motion.
\end{enumerate}





The remainder of this paper is organized as follows.
Section~\ref{sec:method} presents the hybrid sensorless force estimator and its integration with the force-aware whole-body reinforcement learning controller.
Section~\ref{sec:sim_setup} evaluates force-estimation accuracy and force-aware control performance in simulation.
Section~\ref{sec:experiments} presents hardware validation.
Section~\ref{sec:conclusion} concludes the paper.

\section{Method}
\label{sec:method}

\subsection{Problem Formulation and Method Overview}
\label{subsec:problem_overview}

Learning-based locomotion has shown that policies can benefit from estimating or encoding hidden physical variables rather than inferring them only from raw proprioception. 
Examples include learned estimators for base velocity, foot height, and contact probability~\cite{ji2022concurrent}, latent extrinsics for adaptation to terrain, payload, and actuator variations~\cite{kumar2021rma,kumar2022adapting}, and implicit terrain representations for proprioceptive locomotion~\cite{nahrendra2023dreamwaq}. Recent force-control and force-adaptive methods similarly obtain interaction information through learned state estimates or history-based latent representations~\cite{portela2024force, zhi2025unified, facet2025force}. Building on this direction, we use structured interaction-force feedback to support whole-body policy learning for force-controlled loco-manipulation, as conceptually illustrated in Fig.~\ref{fig:interaction_representation}.

In wheeled-legged loco-manipulation, end-effector interaction is coupled with floating-base motion, support contacts, arm dynamics, actuator response, friction, environment contact, and hardware compliance. This coupling is described by the following whole-body dynamics:
\begin{equation}
\mat{M}(\vect{q})\dot{\vect{v}}+\vect{h}(\vect{q},\vect{v})
=
\mat{S}^{\top}\vect{\tau}
+
\mat{J}_{\mathrm{c}}(\vect{q})^{\top}\vect{\lambda}_{\mathrm{c}}
+
\mat{J}_{\mathrm{e}}(\vect{q})^{\top}\vect{w}_{\mathrm{e}}
+
\vect{d}
\label{eq:dynamics}
\end{equation}
where $\vect{q}$ and $\vect{v}$ are the generalized configuration and velocity of the floating-base robot, $\mat{M}$ is the generalized inertia matrix, and $\vect{h}$ collects Coriolis, centrifugal, and gravity terms. The matrix $\mat{S}$ selects the actuated coordinates, and $\vect{\tau}$ contains the commanded actuator torques. The matrices $\mat{J}_{\mathrm{c}}$ and $\mat{J}_{\mathrm{e}}$ are the support-contact and end-effector Jacobians, respectively. The vector $\vect{\lambda}_{\mathrm{c}}$ contains the stacked support-contact forces, $\vect{w}_{\mathrm{e}}$ is the end-effector interaction wrench, and $\vect{d}$ collects unmodeled effects.

Equivalently, the externally induced generalized disturbance is
\begin{equation}
\vect{\xi}
=
\mat{J}_{\mathrm{c}}(\vect{q})^{\top}\vect{\lambda}_{\mathrm{c}}
+
\mat{J}_{\mathrm{e}}(\vect{q})^{\top}\vect{w}_{\mathrm{e}}
+
\vect{d}.
\label{eq:generalized_disturbance}
\end{equation}

Thus, the task-relevant end-effector interaction is not an isolated temporal signal; it is embedded in the same generalized disturbance as support-contact reactions and unmodeled dynamics.

A purely data-driven force estimator can learn useful correlations from proprioceptive history, but does not explicitly reconstruct end-effector force using whole-body dynamics and support-contact constraints. In addition, contact transients and high-frequency force components may be attenuated when they are rare, noisy, or weakly observable in the training data, which is consistent with the averaging tendency of regression losses and the spectral bias of neural networks~\cite{bishop1994mixture,rahaman2019spectral,xu2019frequency}. 
Such estimators may require broad and carefully structured force-labeled data coverage~\cite{osburg2022using,shan2024fine,Shan2024_icra}. When trained mainly in simulation, they may also inherit simulator-specific contact dynamics, actuator models, and force-label biases, which can contribute to sim-to-real degradation in contact-rich tasks~\cite{salvato2021crossing,zhang2023efficient}.

\begin{figure*}[h]
\centering
\includegraphics[width=7in]{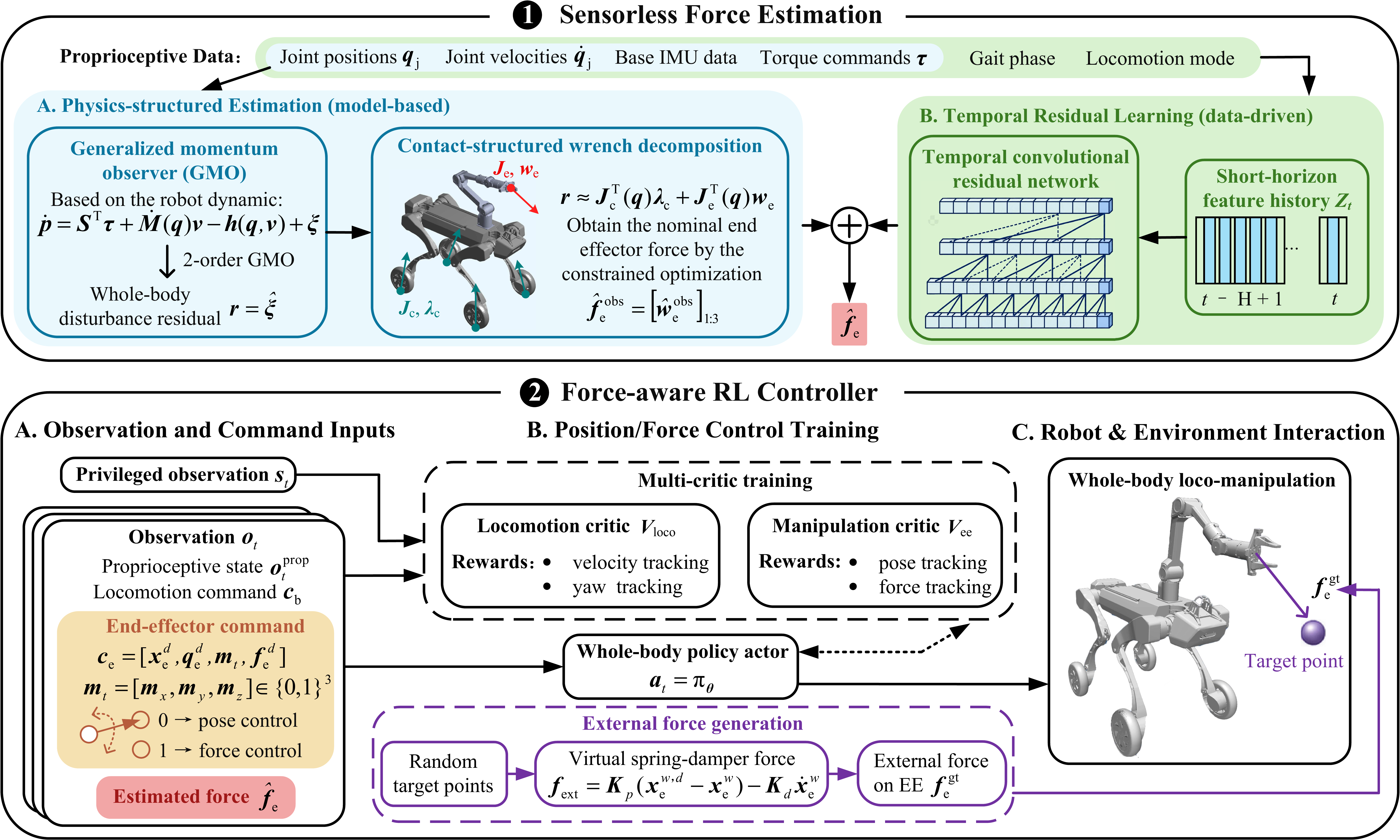}
\vspace{-0.2cm}
\caption{
Overview of the proposed physics-structured interaction representation for force-aware loco-manipulation.
The hybrid sensorless estimator combines physics-structured estimation and temporal residual learning to produce the corrected end-effector force estimate $\hat{\vect{f}}_{\mathrm{e}}$.
This estimate is provided to the RL policy together with proprioceptive observations, locomotion commands, and end-effector commands.
During training, virtual spring--damper interactions and a multi-critic formulation are used to learn both locomotion and manipulation objectives.
The learned policy uses the estimated force feedback for position control, force regulation, and hybrid force/position loco-manipulation without an end-effector force/torque sensor.
}
\vspace{-0.3cm}
\label{fig:force_aware_method}
\end{figure*}

As illustrated in Fig.~\ref{fig:force_aware_method}, we reconstruct a nominal end-effector force from the whole-body disturbance using robot dynamics and contact constraints, then correct its unmodeled bias with temporal residual learning. The corrected force serves as an explicit interaction observation for the whole-body policy, allowing the policy to better coordinate base motion, arm motion, and contact behavior during free-space motion, force-guided interaction, and hybrid force/position loco-manipulation.

\subsection{Generalized Momentum Disturbance Observer}
\label{subsec:observer}

Define the generalized momentum as
\begin{equation}
\vect{p}=\mat{M}(\vect{q})\vect{v} .
\label{eq:momentum}
\end{equation}
Using \eqref{eq:dynamics}, the momentum dynamics can be written as
\begin{equation}
\dot{\vect{p}}
=
\vect{u}(\vect{q},\vect{v},\vect{\tau})+\vect{\xi}
\label{eq:momentum_dynamics}
\end{equation}
where the nominal momentum input is
\begin{equation}
\vect{u}(\vect{q},\vect{v},\vect{\tau})
=
\mat{S}\T\vect{\tau}
+\dot{\mat{M}}(\vect{q})\vect{v}
-\vect{h}(\vect{q},\vect{v}) .
\label{eq:nominal_momentum_input}
\end{equation}

We estimate the whole-body disturbance residual using the following second-order generalized momentum observer:
\begin{align}
\dot{\hat{\vect{p}}}
&=
\vect{u}(\vect{q},\vect{v},\vect{\tau})
+
\hat{\vect{\xi}}
+
\mat{K}_1(\vect{p}-\hat{\vect{p}})
\label{eq:gmo_phat}\\
\dot{\hat{\vect{\xi}}}
&=
\mat{K}_2(\vect{p}-\hat{\vect{p}})
\label{eq:gmo_xihat}
\end{align}
where $\hat{\vect{p}}$ is the estimated generalized momentum, $\hat{\vect{\xi}}$ is the estimated disturbance, and $\mat{K}_1\succ0$, $\mat{K}_2\succ0$ are diagonal observer-gain matrices. The observer residual used for wrench decomposition is
\begin{equation}
\vect{r}=\hat{\vect{\xi}} .
\label{eq:generalized_residual}
\end{equation}

The gains are selected through the second-order parameterization $\mat{K}_1=2\zeta\omega_n\mat{I}$ and $\mat{K}_2=\omega_n^2\mat{I}$, where $\omega_n$ determines the observer bandwidth and $\zeta$ controls damping. Increasing $\omega_n$ improves response speed but increases sensitivity to measurement noise and model mismatch.

The observer converts proprioceptive measurements into a whole-body disturbance representation. However, this representation still mixes support-contact reactions, end-effector interaction, and model-dependent residuals; therefore, it must be further structured before being used as an interaction representation for the policy.

\subsection{Contact-Constrained Wrench Projection}
\label{subsec:decomposition}

The disturbance residual estimated by the momentum observer contains the combined effects of support-contact reactions, end-effector interaction, and residual model mismatch.
Since $\vect{r}$ is an estimated generalized disturbance rather than the true disturbance $\vect{\xi}$, we write the residual used for wrench projection as
\begin{equation}
\vect{r}
=
\mat{B}(\vect{q})\vect{x}
+
\vect{\epsilon}_{\mathrm{r}}
\label{eq:compact_residual_model}
\end{equation}
where $
\mat{B}(\vect{q})
=
\begin{bmatrix}
\mat{J}_{\mathrm{c}}(\vect{q})\T &
\mat{J}_{\mathrm{e}}(\vect{q})\T
\end{bmatrix},
\vect{x}
=
\begin{bmatrix}
\vect{\lambda}_{\mathrm{c}}\T &
\vect{w}_{\mathrm{e}}\T
\end{bmatrix}\T$.
Here, $\vect{\lambda}_{\mathrm{c}}$ denotes the support-contact forces, $\vect{w}_{\mathrm{e}}$ denotes the end-effector wrench, and $\vect{\epsilon}_{\mathrm{r}}$ collects the part of the residual that cannot be explained by the modeled contact and end-effector wrench maps, including unmodeled dynamics and observer estimation error.

A direct least-squares or pseudo-inverse solution is
$\hat{\vect{x}}^{\mathrm{pi}}
=\mat{B}(\vect{q})^\#\vect{r}$,
where $(\cdot)^\#$ denotes the Moore--Penrose pseudo-inverse. This solution explains the observer residual in a minimum-norm sense~\cite{vandam2022collision,lin2026transformer}.
However, in floating-base loco-manipulation, support-contact reactions, end-effector interaction, and model errors are not cleanly separable in generalized coordinates.
As a result, the pseudo-inverse may assign support-contact residuals or model mismatch to the end-effector wrench.
Conversely, hard null-space separation is also unreliable because inertial-parameter errors, Jacobian errors, actuator mismatch, and observer transients can move useful interaction information outside the ideal modeled end-effector wrench subspace.

We therefore estimate the support-contact and end-effector wrench variables using the following soft contact-constrained projection:
\begin{equation}
\begin{aligned}
\hat{\vect{x}}
= \argmin_{\vect{x}} \quad &
\frac{1}{2}
\left\|
\mat{W}_{q}
\left(
\vect{r}-\mat{B}(\vect{q})\vect{x}
\right)
\right\|_2^2
+
\frac{1}{2}\vect{x}\T \mat{R}\vect{x} \\
\mathrm{s.t.}\quad &
\mat{A}_{\mathrm{c}}\vect{x}\leq \vect{b}_{\mathrm{c}} 
\end{aligned}
\label{eq:contact_constrained_qp}
\end{equation}
where $\mat{W}_{q}$ is a weighting matrix for the generalized-disturbance residual. The first term fits the part of the observer residual that is consistent with the modeled wrench maps, while allowing unexplained mismatch to remain in the residual error. The regularization matrix $\mat{R}\succ0$ avoids excessive wrench allocation in redundant directions.
The constraints $\mat{A}_{\mathrm{c}}\vect{x}\leq \vect{b}_{\mathrm{c}}$ encode support-contact feasibility, including unilateral normal contact and friction-cone constraints. Compared with the unconstrained pseudo-inverse, this projection reduces leakage of support-contact disturbances and model mismatch into the estimated end-effector wrench.

It should be noted that the constrained projection does not assume that the support-contact and end-effector wrench contributions are uniquely identifiable from the generalized residual. In a floating-base system, their generalized wrench subspaces may overlap, and model mismatch can further introduce components that are inconsistent with the ideal wrench maps. The role of the contact constraints and regularization is therefore to restrict the admissible wrench allocation and obtain a physically consistent nominal estimate, rather than to enforce an exact decomposition. Remaining ambiguity and motion-dependent modeling errors are subsequently compensated by the temporal residual model. In addition, wheel-ground rolling effects that are not explicitly represented by the force-level contact constraints are treated as part of this residual uncertainty.

After solving \eqref{eq:contact_constrained_qp}, the optimal wrench allocation and nominal end-effector force are given by
\begin{equation}
\hat{\vect{x}}
=
\begin{bmatrix}
\hat{\vect{\lambda}}_{\mathrm{c}}\\
\hat{\vect{w}}_{\mathrm{e}}^{\mathrm{obs}}
\end{bmatrix},
\qquad
\hat{\vect{f}}_{\mathrm{e}}^{\mathrm{obs}}
=
\left[
\hat{\vect{w}}_{\mathrm{e}}^{\mathrm{obs}}
\right]_{1:3},
\label{eq:nominal_force_estimate}
\end{equation}
where $\hat{\vect{\lambda}}_{\mathrm{c}}$ contains the estimated support-contact forces and $\hat{\vect{w}}_{\mathrm{e}}^{\mathrm{obs}}$ is the nominal end-effector wrench.
The operator $[\cdot]_{1:3}$ extracts the first three wrench components, corresponding to the interaction force.
The superscript $\mathrm{obs}$ identifies the estimate from the observer--projection stage before temporal residual correction.

\subsection{Temporal Residual Learning}
\label{subsec:residual_learning}

Although the contact-constrained wrench projection provides a contact-consistent nominal force estimate, residual errors can remain because the observer residual is still affected by unmodeled dynamics, actuator mismatch, contact uncertainty, sensing error, and observer transients.
These errors are difficult to model analytically, but they are correlated with recent robot motion, support-contact evolution, gait phase, and locomotion mode. We therefore use temporal residual learning to compensate only the remaining bias of the nominal estimate.

The residual learning target is defined as
\begin{equation}
\Delta \vect{f}_{\mathrm{e},t}^{\star}
=
\vect{f}_{\mathrm{e},t}^{\mathrm{gt}}
-
\hat{\vect{f}}_{\mathrm{e},t}^{\mathrm{obs}}
\label{eq:residual_target}
\end{equation}
where $\vect{f}_{\mathrm{e},t}^{\mathrm{gt}}$ is the ground-truth interaction force and $\hat{\vect{f}}_{\mathrm{e},t}^{\mathrm{obs}}$ is the nominal estimate from the contact-constrained projection.

At each time step, the temporal model receives the feature vector
\begin{equation}
\vect{z}_t
=
\left[
\vect{x}_t^{\mathrm{dyn}},
\vect{x}_t^{\mathrm{base}},
\vect{x}_t^{\mathrm{gait}},
\vect{x}_t^{\mathrm{mode}}
\right]
\label{eq:feature_decomp}
\end{equation}
where $\vect{x}_t^{\mathrm{dyn}}$ contains joint positions, joint velocities, and torque commands; $\vect{x}_t^{\mathrm{base}}$ contains base orientation and angular velocity; $\vect{x}_t^{\mathrm{gait}}$ encodes gait phase; and $\vect{x}_t^{\mathrm{mode}}$ indicates the locomotion mode.

Given a history window $\mathcal{Z}_t = \left\{\vect{z}_{t-H+1},\ldots,\vect{z}_t \right\}$, a causal temporal convolutional network predicts the residual force correction
\begin{equation}
\Delta \hat{\vect{f}}_{\mathrm{e},t}^{\mathrm{res}}
=
\Phi_\theta(\mathcal{Z}_t)
\label{eq:tcn_predictor}
\end{equation}
where $\Phi_\theta$ denotes the causal temporal convolutional network (TCN) with trainable parameters $\theta$.

The final hybrid force estimate is
\begin{equation}
\hat{\vect{f}}_{\mathrm{e},t}
=
\hat{\vect{f}}_{\mathrm{e},t}^{\mathrm{obs}}
+
\Delta \hat{\vect{f}}_{\mathrm{e},t}^{\mathrm{res}} .
\label{eq:hybrid_estimate}
\end{equation}

We use a causal TCN with a history length of $H=20$ time steps and five residual temporal blocks with dilation factors $\{1,2,4,8,16\}$.
The channel width is $64$, the kernel size is $3$, and an MLP head with a $128$-unit hidden layer maps the latest temporal feature to the
three-dimensional residual force correction. A dropout rate of $0.1$ is used during training. The network is trained with a Smooth-$\ell_1$ loss:
\begin{equation}
\mathcal{L}_{\mathrm{res}}
=
\frac{1}{N}
\sum_{i=1}^{N}
\rho_{\delta}
\left(
\Phi_{\theta}(\mathcal{Z}_i)
-
\Delta \vect{f}_{\mathrm{e},i}^{\star}
\right)
\label{eq:loss}
\end{equation}
where $N$ is the number of supervised samples included in the loss, and $\rho_\delta$ denotes the Smooth-$\ell_1$ loss with transition parameter $\delta>0$.

Training data are collected offline at $50~\mathrm{Hz}$ under randomized base and end-effector commands and randomly directed end-effector forces of varying magnitude. Rollouts are separated into training, validation, and held-out test subsets before temporal windows are constructed, and normalization statistics are computed from the training subset only.

\subsection{Force-Aware Whole-Body Reinforcement Learning}
\label{subsec:force_aware_rl}

The corrected force estimate is integrated into the mode-conditioned whole-body reinforcement learning controller shown in Fig.~\ref{fig:force_aware_method}. The purpose of the controller is not merely to track end-effector pose, but to regulate physical interaction using explicit estimated-force feedback.


At each control step, the policy receives the observation
\begin{equation}
\begin{aligned}
\vect{o}_t =
\big[
&\vect{\omega}_{\mathrm{b}},\;
\vect{g}_{\mathrm{b}},\;
\vect{c}_{\mathrm{b}},\;
\vect{c}_{\mathrm{e}},\;
\vect{q}_{\ell},\;
\vect{q}_{\mathrm{a}},\;
\dot{\vect{q}}_{\mathrm{j}},\;
\hat{\vect{f}}_{\mathrm{e}},\;
\vect{a}_{t-1},\;
\vect{\phi}_{\mathrm{g}}
\big]
\end{aligned}
\label{eq:rl_observation}
\end{equation}
where $\vect{\omega}_{\mathrm{b}}$ is the base angular velocity, $\vect{g}_{\mathrm{b}}$ is the projected gravity vector, $\vect{c}_{\mathrm{b}}$ is the locomotion command, $\vect{c}_{\mathrm{e}}$ is the end-effector task command, $\vect{q}_{\ell}$ and $\vect{q}_{\mathrm{a}}$ are the leg and arm joint positions, $\dot{\vect{q}}_{\mathrm{j}}$ contains the actuated joint velocities, $\vect{a}_{t-1}$ is the previous action, and $\vect{\phi}_{\mathrm{g}}$ denotes the gait phase. The force-feedback term $\hat{\vect{f}}_{\mathrm{e}}$ is provided by the proposed hybrid sensorless estimator. All Cartesian end-effector position commands, force commands, and estimated interaction forces are expressed in a body-aligned world frame, whose horizontal axes follow the robot heading while the vertical axis remains aligned with the world gravity direction.

The end-effector command is defined as
\begin{equation}
\vect{c}_{\mathrm{e}}
=
\left[
\vect{x}_{\mathrm{e}}^d,\;
\vect{q}_{\mathrm{e}}^d,\;
\vect{m}_t,\;
\vect{f}_{\mathrm{e}}^d
\right]
\label{eq:ee_command_force_rl}
\end{equation}
where $\vect{x}_{\mathrm{e}}^d\in\R^{3}$ is the desired end-effector position, $\vect{q}_{\mathrm{e}}^d\in\R^{4}$ is the unit quaternion representing the desired end-effector orientation, $\vect{f}_{\mathrm{e}}^d\in\R^{3}$ is the desired interaction force, and
$ \vect{m}_t = \begin{bmatrix} m_{x,t} & m_{y,t} & m_{z,t} \end{bmatrix}\T \in \{0,1\}^{3}$ is the axis-wise force/position selector, where for axis $i\in\{x,y,z\}$, $m_{i,t}=0$ indicates position tracking while $m_{i,t}=1$ indicates force regulation.

The selector $ \vect{m}_t$ allows a single policy to represent multiple manipulation modes. When all mask entries are zero, the policy performs free-space end-effector position tracking. When all entries are one, the policy performs pure force regulation. Mixed masks produce hybrid force/position control, such as tangential position tracking with normal-force regulation during wiping.

The policy outputs a 22-dimensional whole-body action consisting of desired leg and arm joint positions and desired wheel velocities. The leg and arm commands are tracked by joint-space PD controllers, while the wheels are controlled in velocity mode. The policy is executed at $50~\mathrm{Hz}$, with gains of $K_p=40$, $K_d=2$ for the leg joints, $K_d=3$ for wheel velocity control, and $K_p=15$, $K_d=1$ for the arm joints.

The manipulation reward is computed according to the active force/position assignment. Let
\begin{equation}
\bar{\vect{m}}_t = \vect{1}_3 - \vect{m}_t 
\label{eq:position_mode_mask}
\end{equation}
where $\vect{1}_3=[1,1,1]\T$, and let $\odot$ denote element-wise multiplication. 

The position-tracking reward and force-regulation reward are
\begin{equation}
r_{\mathrm{pos}}
=
\exp
\left(
-
\frac{
\left\|
\bar{\vect{m}}_t \odot
\left(\vect{x}_{\mathrm{e}}-\vect{x}_{\mathrm{e}}^d\right)
\right\|_1
}{\sigma_x}
-
\frac{\|\vect{e}_{R}\|_1}{\sigma_R}
\right)
\label{eq:axiswise_pose_tracking_reward}
\end{equation} 

\begin{equation}
r_{\mathrm{force}}
=
\exp
\left(
-
\frac{
\left\|
\vect{m}_t \odot
\left({\vect{f}}_{\mathrm{e}}-\vect{f}_{\mathrm{e}}^d\right)
\right\|_2^2
}{\sigma_f}
\right)
\label{eq:axiswise_force_tracking_reward}
\end{equation}
where $\vect{x}_{\mathrm{e}}$ is the current end-effector position, expressed in the same frame as $\vect{x}_{\mathrm{e}}^d$, and $\vect{e}_{R}$ is the end-effector orientation error. $\sigma_x$, $\sigma_R$, and $\sigma_f$ are tracking tolerances. ${\vect{f}}_{\mathrm{e}}$ denotes the simulator ground-truth end-effector interaction force. The ground-truth interaction force is used only for reward computation during policy training and is not included in the policy observation. The policy instead receives the sensorless estimate $\hat{\vect{f}}_{\mathrm{e}}$ as its force observation.

The end-effector reward is then written as
\begin{equation}
r_t^{\mathrm{ee}}
=
w_{\mathrm{pos}} r_{\mathrm{pos}}
+
w_{\mathrm{force}} r_{\mathrm{force}}
\label{eq:axiswise_ee_reward}
\end{equation}
where $w_{\mathrm{pos}}$ and $w_{\mathrm{force}}$ balance the position-tracking and force-regulation objectives.

To train force-regulation behavior without constructing a separate object model, we generate end-effector interaction forces in simulation using a virtual spring-damper model:
\begin{equation}
\vect{f}_{\mathrm{ext}}
=
\mat{K}_{p}
\left(
\vect{x}_{\mathrm{e}}^{w,d}-\vect{x}_{\mathrm{e}}^w
\right)
-
\mat{K}_{d} \dot{\vect{x}}_{\mathrm{e}}^w 
\label{eq:virtual_pd_force}
\end{equation}
where $\vect{x}_{\mathrm{e}}^w$ and $\dot{\vect{x}}_{\mathrm{e}}^w$ are the current end-effector position and velocity in the world frame, $\vect{x}_{\mathrm{e}}^{w,d}$ is a sampled target point transformed into the world frame, and $\mat{K}_{p},\mat{K}_{d}$ are virtual stiffness and damping gain matrices. The commanded force components used for policy training are sampled within $[-10,10]~\mathrm{N}$.

The total reward is decomposed into locomotion and manipulation components:
\begin{equation}
r_t
=
r_t^{\mathrm{loco}}
+
r_t^{\mathrm{ee}}
\label{eq:total_rl_reward}
\end{equation}
where $r_t^{\mathrm{loco}}$ contains base velocity tracking, yaw tracking, gait regularization, stability, and torque-related terms, while $r_t^{\mathrm{ee}}$ contains the axis-wise end-effector position and force terms. The policy is trained using PPO with a multi-critic actor--critic architecture. The actor is an MLP with hidden dimensions $[512,256,128]$ and ELU activations. The critic uses a shared MLP trunk with the same hidden dimensions and two scalar value heads corresponding to the locomotion and end-effector objectives.

Through this design, the hybrid sensorless estimate serves as an explicit feedback variable, allowing the policy to directly use estimated interaction force for force regulation and hybrid force/position loco-manipulation rather than infer contact force from proprioception alone.

\section{Simulation Experiments}
\label{sec:sim_setup}

This section evaluates whether the proposed structured interaction representation improves both sensorless force estimation and force-aware control behavior.

\begin{figure*}[t]
\centering
\includegraphics[width=1.0\textwidth]{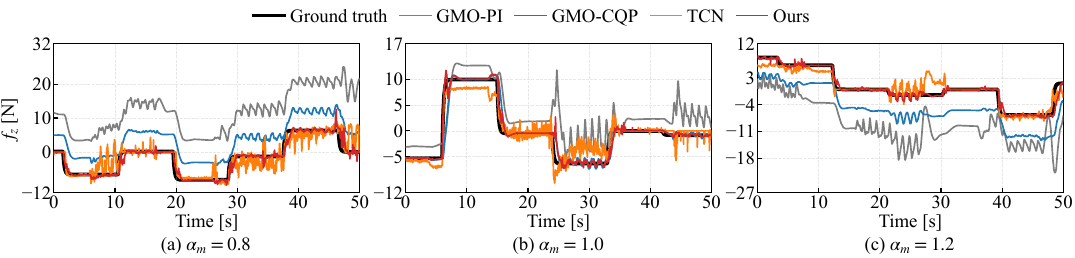}
\vspace{-0.6cm}
\caption{
Representative $f_z$ force-estimation results under inertial-parameter mismatch.
The inertial parameters used by the generalized momentum observer and wrench projection are kept fixed, while the simulated robot inertial parameters are scaled by $\alpha_m=0.8$, $1.0$, and $1.2$.
The compared methods are GMO-PI, GMO-CQP, TCN, and the proposed estimator (Ours).
}
\label{fig:force_est_mismatch}
\vspace{-0.3cm}
\end{figure*}

\subsection{Simulation Protocol}
\label{subsec:sim_setup}

We conduct two simulation evaluations. The first evaluates the accuracy of the proposed hybrid sensorless force estimator. The robot executes randomized base and arm commands while randomized external forces are applied at the end effector, covering diverse whole-body motion and interaction conditions. The networks are trained offline and kept fixed during evaluation on separate MuJoCo rollouts. For each inertial setting, ten independent rollouts are evaluated; all estimators receive identical trajectories and applied-force sequences, and the simulator ground-truth force is used only to compute errors.
Results are reported as mean $\pm$ sample standard deviation across rollouts.

The second evaluation compares the force-control and manipulation performance of the proposed method with two reproduced learning-based baselines. We use a lateral box-dragging task, where the robot must generate tangential dragging force while maintaining a desired normal pressing force. All compared controllers are evaluated with the same robot model, object model, contact parameters, initial condition, and force-command sequence. Each method is evaluated in ten trials. Force RMSEs are reported as mean $\pm$ sample standard deviation, while box-motion extrema are reported as trial means. The simulator ground-truth contact force is used only for evaluation and is not provided to the policy.

\subsection{Estimator Baselines and Ablations}
\label{subsec:baselines}

We compare the proposed hybrid estimator with three estimator baselines:

\begin{itemize}
    \item \textbf{GMO-PI}: the generalized momentum observer followed by a stacked-Jacobian pseudo-inverse decomposition. This baseline evaluates an unconstrained analytical decomposition of the observer residual into support-contact and end-effector wrench components.

    \item \textbf{GMO-CQP}: the generalized momentum observer followed by the contact-constrained quadratic-programming decomposition. This baseline evaluates the benefit of imposing contact feasibility during residual decomposition, without using learned temporal compensation.

    \item \textbf{TCN}: a learning-only force estimator that directly predicts the end-effector force from proprioceptive history, without using the model-based nominal estimate.

    \item \textbf{Ours}: the proposed hybrid estimator, which adds temporal residual compensation to the contact-constrained nominal estimate.
\end{itemize}

These baselines isolate the contribution of each component. The comparison between GMO-PI and GMO-CQP evaluates whether contact feasibility reduces the ambiguity between support-contact forces and end-effector interaction. The comparison between GMO-CQP and our method evaluates whether temporal residual learning compensates the remaining motion-dependent bias. The standalone TCN uses the same temporal architecture and model capacity as the residual TCN in the proposed method; therefore, the comparison isolates the difference between residual correction and direct force prediction.

\subsection{Force-Estimation Results}
\label{subsec:force_results}

We evaluate force-estimation performance under three estimator-model settings: $\alpha_m=0.8$, $\alpha_m=1.0$, and $\alpha_m=1.2$. In this evaluation, the inertial parameters used by the generalized momentum observer and wrench projection are kept fixed, while the inertial parameters of the simulated robot are scaled as
$m_i^{\mathrm{sim}}=\alpha_m m_i^{\mathrm{est}}$ and
$I_i^{\mathrm{sim}}=\alpha_m I_i^{\mathrm{est}}$,
where $m_i^{\mathrm{sim}}$ and $I_i^{\mathrm{sim}}$ denote the inertial parameters of the simulated robot model, and $m_i^{\mathrm{est}}$ and $I_i^{\mathrm{est}}$ denote those used by the estimator. Thus, $\alpha_m=1.0$ corresponds to the matched-model case, while $\alpha_m=0.8$ and $\alpha_m=1.2$ correspond to simulated robots that are lighter and heavier than the estimator model, respectively. For both the standalone TCN and the proposed residual estimator, one network is trained using the pooled training rollouts from all settings. The selected network parameters and normalization statistics are then kept fixed when evaluating held-out rollouts from each setting. This setting evaluates the force-estimation performance of the proposed hybrid formulation under different levels of mismatch between the model-based estimator and the simulated robot dynamics.



\begin{table}[htp]
\centering
\caption{Force-estimation RMSE under inertial-parameter mismatch.}
\label{tab:force_estimation_mismatch}
\begin{minipage}{\columnwidth}
\centering
\footnotesize
\setlength{\tabcolsep}{2.0pt}
\renewcommand{\arraystretch}{1.05}

\begin{tabular*}{\columnwidth}
{@{\extracolsep{\fill}}clcccc@{}}
\hline
$\alpha_m$ & Method & $e_x$ & $e_y$ & $e_z$ & $e_{\mathrm{vec}}$ \\
\hline
0.8 & GMO-PI
& 5.89 & 5.76 & 14.76 & $16.97\pm0.91$ \\
0.8 & GMO-CQP
& 1.33 & 1.25 & 5.69 & $5.99\pm0.13$ \\
0.8 & TCN
& 2.90 & 3.24 & 3.10 & $5.41\pm0.33$ \\
0.8 & Ours
& \textbf{1.17} & \textbf{1.11} & \textbf{1.17}
& $\bm{2.02\pm0.31}$ \\
\hline
1.0 & GMO-PI
& 5.37 & 5.67 & 3.76 & $8.72\pm1.11$ \\
1.0 & GMO-CQP
& 1.19 & 1.19 & 1.14 & $2.06\pm0.39$ \\
1.0 & TCN
& 2.20 & 2.50 & 2.05 & $3.96\pm0.87$ \\
1.0 & Ours
& \textbf{0.98} & \textbf{1.04} & \textbf{0.87}
& $\bm{1.68\pm0.33}$ \\
\hline
1.2 & GMO-PI
& 6.80 & 6.97 & 13.36 & $16.59\pm1.25$ \\
1.2 & GMO-CQP
& 1.28 & 1.32 & 5.36 & $5.68\pm0.10$ \\
1.2 & TCN
& 3.90 & 3.04 & 3.77 & $6.28\pm0.96$ \\
1.2 & Ours
& \textbf{1.10} & \textbf{1.03} & \textbf{0.99}
& $\bm{1.82\pm0.35}$ \\
\hline
\end{tabular*}

\vspace{0.05cm}
\raggedright
\footnotesize
Note: $e_x$, $e_y$, and $e_z$ denote axis-wise force-estimation RMSEs, and $e_{\mathrm{vec}}$ denotes the RMSE of the Euclidean force-error magnitude. Axis-wise entries are rollout means; $e_{\mathrm{vec}}$ is reported
as mean $\pm$ sample standard deviation over ten rollouts.
All values are in newtons.
\end{minipage}
\vspace{-0.2cm}
\end{table}

Fig.~\ref{fig:force_est_mismatch} shows representative time-series results of the vertical interaction force $f_z$ under the three estimator-model settings.
GMO-PI captures the general trend of the interaction force but exhibits noticeable bias and fluctuation, especially under inertial mismatch.
This behavior is expected because the pseudo-inverse decomposition distributes the observer residual in a least-squares sense.
As a result, model errors and support-contact residuals can leak into the estimated end-effector wrench.

GMO-CQP reduces this ambiguity by enforcing contact-feasibility constraints and regularizing the wrench allocation.
Compared with GMO-PI, it produces a more physically consistent nominal estimate and substantially reduces large force-estimation errors. However, GMO-CQP can still contain residual bias because the observer residual is affected by inertial mismatch, rolling dynamics, floating-base motion, friction, and support-contact uncertainty. These effects are difficult to remove using model-based decomposition alone.

Table~\ref{tab:force_estimation_mismatch} shows that GMO-CQP reduces the mean vector RMSE relative to GMO-PI in all three settings. Adding temporal residual compensation yields the lowest error across the evaluated settings, with mean vector RMSEs of $1.68$--$2.02$~N. Relative to the standalone TCN, the proposed estimator reduces this error by $57.6$--$71.0\%$. Its improvement over GMO-CQP is greater in the two mismatched-model settings than in the matched-model setting, indicating the value of residual compensation when the nominal model is inaccurate.



These results indicate that, for the considered force-estimation problem, learning is more effective when used to compensate the residual error of a physically constrained nominal estimate than when used to infer the complete interaction force directly from proprioceptive history. Such a residual formulation is attractive for real robots, where collecting broad force-labeled datasets under diverse contacts, locomotion modes, and hardware conditions is costly and difficult. Overall, the results support the hybrid design: the model-based projection provides a contact-consistent nominal estimate, while temporal residual learning compensates the remaining motion-dependent bias.

\subsection{Force Control Comparison}
\label{subsec:force_policy_comparison}

We further evaluate the force-control and manipulation performance of the proposed method in a box-dragging task. The robot is required to move the box along the $y$ direction while maintaining a normal pressing force of $20~\mathrm{N}$ along the $z$ direction. This task evaluates whether force tracking produces the desired object motion.
\begin{figure}[htp]
    \centering
    \includegraphics[width=3.5in]{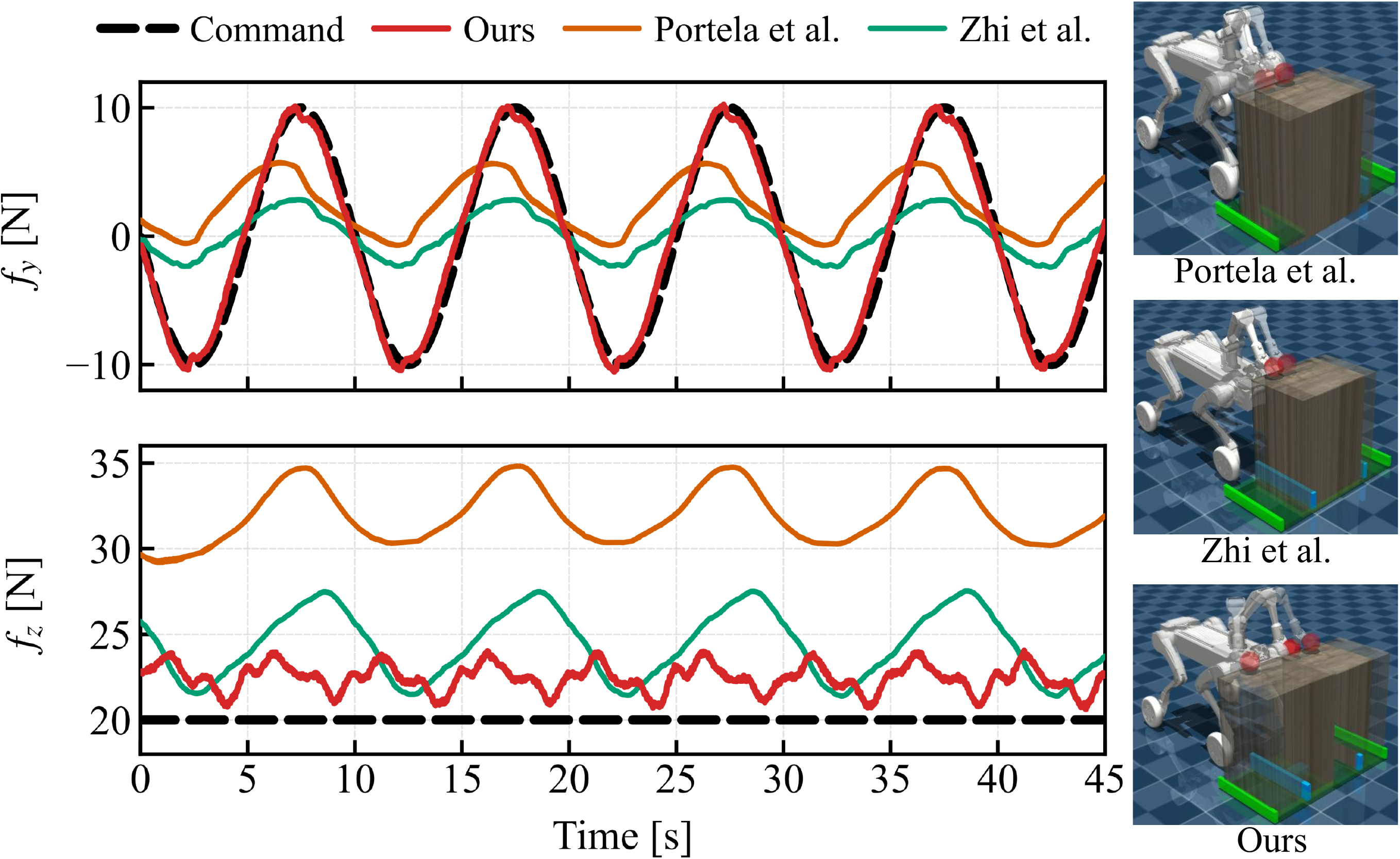}
    \vspace{-0.55cm}
    \caption{
    Pure force-control comparison in the lateral box-dragging task. The snapshots on the right side show the resulting box motion, where the proposed method moves the box closer to the expected lateral displacement range. The left plots show the ground-truth tangential force $f_y$ and normal force $f_z$ generated by each controller. Compared with the reproduced baselines~\cite{portela2024force, zhi2025unified}, the proposed method tracks the commanded $f_y$ more accurately while maintaining $f_z$ near the desired $20~\mathrm{N}$ level.
    }
    \label{fig:drag_box_force_compare}
    \vspace{-0.2cm}
\end{figure}

\begin{table}[htp]
\vspace{-0.3cm}
\centering
\caption{Box-dragging performance over ten trials.}
\label{tab:drag_box_force_comparison}
\begin{minipage}{\columnwidth}
\centering
\footnotesize
\setlength{\tabcolsep}{1.5pt}
\renewcommand{\arraystretch}{1.05}

\begin{tabular*}{\columnwidth}
{@{\extracolsep{\fill}}lccc@{}}
\hline
Method & $e_y$ [N] & $e_z$ [N] & $y_{\mathrm{box}}$ [m] \\
\hline
Portela~\cite{portela2024force}
& $6.21\pm0.19$
& $12.69\pm0.80$
& $[-0.199,\,0.007]$ \\

Zhi~\cite{zhi2025unified}
& $5.42\pm0.03$
& $5.44\pm0.06$
& $[-0.049,\,0.039]$ \\

Ours
& $\bm{0.98\pm0.04}$
& $\bm{3.98\pm0.04}$
& $\bm{[-0.205,\,0.220]}$ \\
\hline
\end{tabular*}

\vspace{0.05cm}
\raggedright
\footnotesize
Note: $e_y$ and $e_z$ denote force-tracking RMSEs relative to the commanded forces. Force RMSEs are reported as mean $\pm$ sample standard deviation.
The final column gives the mean box-motion extrema across ten trials;
the desired range is approximately $[-0.2,0.2]~\mathrm{m}$.
\end{minipage}
\vspace{-0.2cm}
\end{table}

We compare the proposed method with two learning-based force-control baselines~\cite{portela2024force, zhi2025unified}, reproduced using their official open-source implementations and evaluated under the common conditions described above. For both reproduced baselines, the commanded force components are sampled within $[-30,30]~\mathrm{N}$ during training. The Portela et al. baseline~\cite{portela2024force} supplies observation history and learned state estimates, including gripper force, to the actor. The Zhi et al. baseline~\cite{zhi2025unified} supplies shared history features to the actor and a state-estimation head, and learns force/position behavior through impedance-derived motion-tracking objectives.


Fig.~\ref{fig:drag_box_force_compare} shows the ground-truth
contact forces generated by the compared controllers. The proposed
controller follows the commanded lateral force with smaller phase
and amplitude errors. The reproduced Portela et al. baseline captures the periodic lateral-force trend, but its response is attenuated and its normal force exceeds the desired level. Although its actor receives an explicit force estimate, this estimate is learned from observation history without an explicit dynamics-based reconstruction of end-effector interaction. Errors in the learned force feedback under changing whole-body motion and contact conditions may contribute to the observed force-regulation errors.

The reproduced Zhi et al. baseline avoids severe over-pressing, but its lateral-force response remains conservative. This behavior is consistent with an indirect force-control mechanism based on motion compensation: when the contact stiffness and sliding dynamics differ from the assumed impedance behavior, the resulting force response can become weak for dynamic force tracking. In contrast, the proposed policy receives an explicit sensorless force estimate and is trained with an axis-wise force objective. This allows the controller to achieve both dynamic tangential-force tracking and stable normal-force regulation during sliding contact.

\begin{figure*}[t]
\centering
\includegraphics[width=1.0\textwidth]{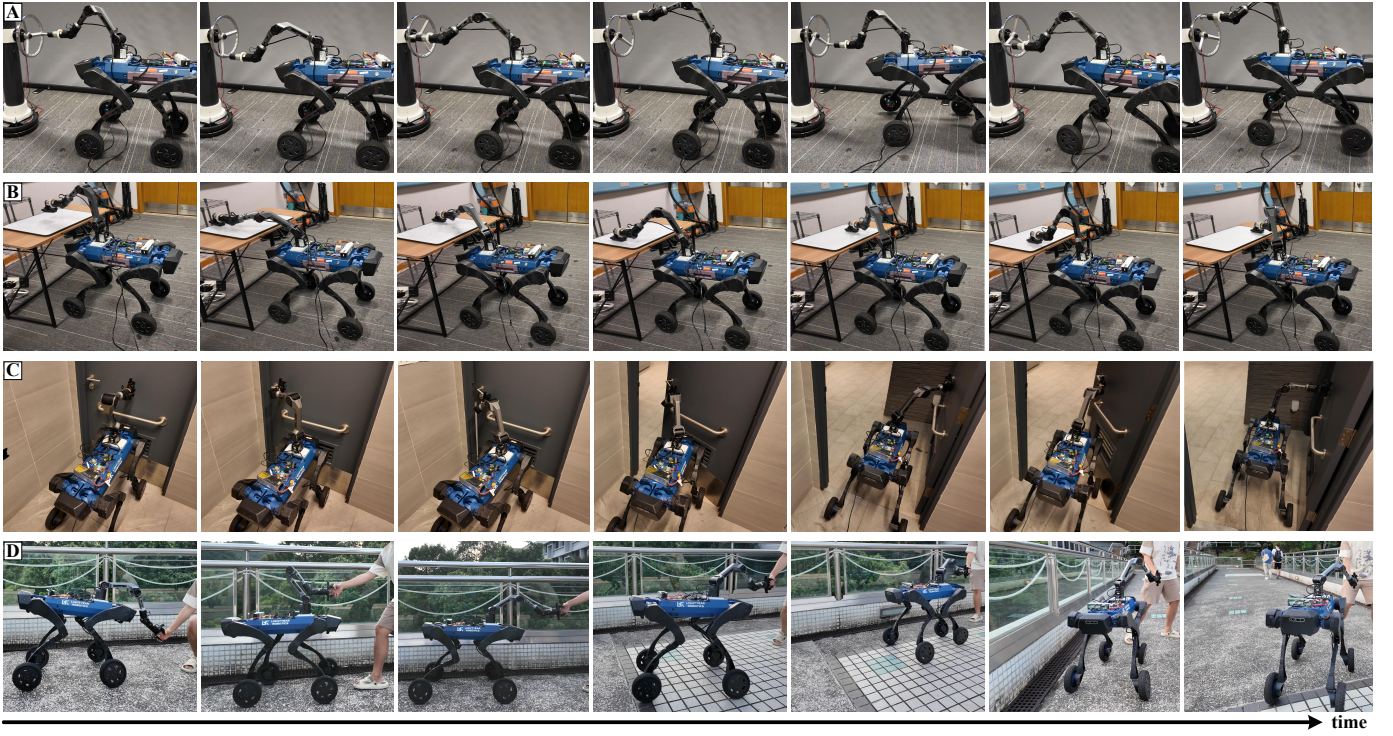}
\vspace{-0.6cm}
\caption{
Representative temporal sequences of the hardware experiments.
(A) Valve rotation while coordinating arm interaction and whole-body motion.
(B) Hybrid force/position wiping with tangential end-effector motion and sustained normal contact.
(C) Force-guided opening of a spring-loaded door through
coordinated manipulation and base motion.
(D) Zero-force human-guided motion, in which external interaction is used to guide the whole-body robot motion.
Images in each row are ordered from left to right in time.
}
\label{fig:hardware_experiment_sequences}
\vspace{-0.3cm}
\end{figure*}

Table~\ref{tab:drag_box_force_comparison} confirms the force and motion trends in Fig.~\ref{fig:drag_box_force_compare}. The proposed method achieves the lowest mean force-tracking errors, with an $f_y$ RMSE of $0.98~\mathrm{N}$ and an $f_z$ RMSE of $3.98~\mathrm{N}$, outperforming both reproduced baselines. Moreover, the proposed method produces the expected bidirectional box motion, reaching mean extrema of $[-0.205,0.220]~\mathrm{m}$, which closely match the desired range of $[-0.2,0.2]~\mathrm{m}$. In contrast, Portela et al. and Zhi et al. reach only $[-0.199,0.007]~\mathrm{m}$ and $[-0.049,0.039]~\mathrm{m}$, respectively. These results show that the proposed controller achieves more accurate force regulation and more effective box manipulation than the reproduced baselines in the evaluated task.

\section{Hardware Experiments}
\label{sec:experiments}

We evaluate the proposed force-aware whole-body RL controller on our wheeled-legged platform in four contact-rich tasks, as summarized in Fig.~\ref{fig:hardware_experiment_sequences}: valve rotation, hybrid force/position wiping, force-guided door opening, and zero-force human-guided motion. These experiments examine sustained force regulation, simultaneous force/position control, coordinated interaction with an articulated environment, and compliant response to externally applied forces.

Component-level estimator comparisons are conducted in simulation, where identical trajectories, applied forces, and controlled model mismatch enable direct attribution. Hardware experiments instead evaluate the frozen estimator--controller system end to end under real motion and contact variation. Valve rotation and wiping use an end-effector F/T sensor only as an external quantitative reference for force estimation and regulation, never as input to the estimator or policy; door opening and human-guided motion provide complementary system-level demonstrations. For hardware deployment, the observer and QP run at $500~\mathrm{Hz}$ with $\mat{K}_1=6\mat{I}$, $\mat{K}_2=20\mat{I}$; the residual network and control policy run at $50~\mathrm{Hz}$.

Across the hardware experiments, we additionally evaluate whether the policy generalizes beyond the force-command range encountered during training. Although the commanded force components are limited to $[-10,10]~\mathrm{N}$ during policy training, the hardware tasks use commands of approximately $18~\mathrm{N}$ during valve rotation, $15~\mathrm{N}$ during board wiping, and up to approximately $30~\mathrm{N}$ during door opening.

\subsection{Valve-Rotation Experiment}
\label{subsec:valve_rotation}

The first experiment evaluates force-aware loco-manipulation in a valve-rotation task. The wheeled-legged robot uses the mounted 6-DoF arm to establish contact with the valve and generate the Cartesian interaction force required for rotation. Compared with a preplanned position-controlled strategy, which typically relies on knowledge of the valve pose, rotation axis, and end-effector trajectory, the proposed force-aware controller rotates the valve through regulated interaction force. This reduces the dependence of task execution on an accurately specified valve trajectory and interaction geometry.

\begin{figure}[htp]
\centering
\includegraphics[width=3.4in]{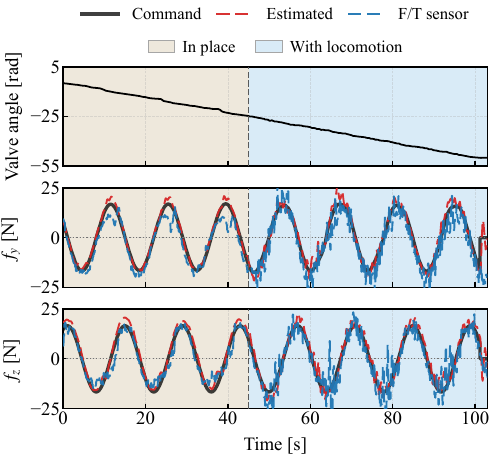}
\vspace{-0.4cm}
\caption{
Representative valve-rotation trial during in-place manipulation and manipulation during locomotion. The top panel shows the valve rotation angle, while the middle and bottom panels compare the commanded, estimated, and measured interaction forces along the $y$ and $z$ axes. The dashed vertical line marks the transition from in-place manipulation to manipulation during locomotion.
}
\label{fig:valve_rotation_force}
\vspace{-0.2cm}
\end{figure}

To evaluate force regulation under different locomotion conditions, the valve-rotation experiment comprises two phases. In the first phase, the robot rotates the valve in place while allowing whole-body posture adjustments. In the second phase, the robot continues rotating the valve during locomotion. The second phase introduces additional floating-base motion, support-contact variation, and hardware compliance, making force estimation and force regulation more challenging.


Fig.~\ref{fig:valve_rotation_force} shows the valve rotation angle and the corresponding interaction-force response. During in-place manipulation, the estimated and measured forces exhibit consistent periodic trends in the $y$ and $z$ directions, corresponding to the cyclic interaction required for valve rotation.



When locomotion begins, the measured force contains larger
fluctuations due to whole-body motion, contact variation, valve friction, and hardware compliance. Despite these disturbances, the sensorless estimate preserves the dominant force trend used by the policy. In both phases, the onboard manipulator continuously rotates the valve through force-guided contact, demonstrating task completion without a predefined valve-handle trajectory.

For the representative trial in Fig.~\ref{fig:valve_rotation_force},
the force-estimation RMSEs during the in-place manipulation phase are $4.91~\mathrm{N}$ and $4.12~\mathrm{N}$ along the $y$ and $z$ axes,
respectively, while the corresponding force-control RMSEs are
$3.99~\mathrm{N}$ and $3.15~\mathrm{N}$. During the locomotion phase, the estimation RMSEs increase to $6.47~\mathrm{N}$ and
$6.38~\mathrm{N}$, while the force-control RMSEs are
$5.82~\mathrm{N}$ and $5.42~\mathrm{N}$. The increased errors reflect the additional whole-body motion and contact variation, while the robot continues the valve-rotation task.


\subsection{Hybrid Force/Position Wiping Experiment}
\label{subsec:wiping_experiment}

The second experiment evaluates hybrid force/position control by tracking tangential wiping motion while regulating the normal contact force of a wiping tool against a board.

During the wiping experiment, the robot heading is maintained such that the board task axes are aligned with the body-aligned world frame used by the controller. The \(x\)-\(y\) plane therefore corresponds to the board-tangential directions, while the \(z\) axis corresponds to the surface-normal direction. The controller tracks the desired wiping motion in the tangential plane and regulates the normal interaction force to the commanded value $f_z^d=15~\mathrm{N}$.

\begin{figure}[htp]
\centering
\includegraphics[width=3.4 in]{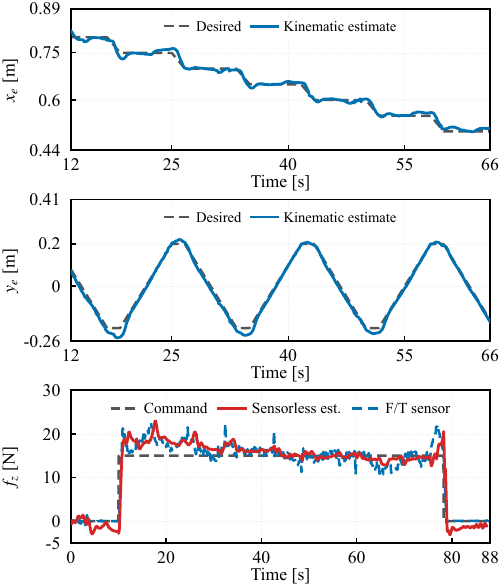}
\vspace{-0.2cm}
\caption{
Representative hybrid force/position wiping trial. The desired and kinematically estimated end-effector positions along the board-tangential $x$ and $y$ directions are shown in the top and middle panels, respectively. The bottom panel shows the commanded normal force, the proposed sensorless force estimate, and the external F/T sensor measurement.}
\label{fig:wiping_experiment}
\vspace{-0.2cm}
\end{figure}

Fig.~\ref{fig:wiping_experiment} shows the resulting hybrid force/position behavior. In the tangential directions, the kinematically estimated end-effector trajectory follows the desired wiping pattern accurately. As the normal contact is established, the commanded force rises to approximately $15~\mathrm{N}$, and the estimated force follows the desired level during sustained sliding contact. The measured force contains fluctuations caused by surface friction, tool-board contact variation, and mechanical compliance. Nevertheless, the sensorless estimate follows the main trend of the measured force and provides stable feedback for the force-aware policy.

For the representative trial in Fig.~\ref{fig:wiping_experiment}, the planar position RMSE computed from the kinematically reconstructed end-effector motion is $e_{xy}=2.90~\mathrm{cm}$. For normal-force regulation, the RMSE between the measured force and the $15$~N command is $2.13$~N, while the RMSE between the sensorless estimate and the F/T measurement is $2.74$~N. These results demonstrate the axis-wise hybrid-control capability of the proposed approach: the policy simultaneously tracks tangential end-effector motion and regulates the normal interaction force during sustained contact.

\subsection{Force-Guided Opening of a Spring-Loaded Door}
\label{subsec:door_opening}

The third hardware experiment evaluates the proposed force-aware
controller in a multi-stage task involving a spring-loaded door. The spring mechanism produces a restoring torque that opposes opening, requiring sustained interaction to overcome the closing load. The robot must establish contact with the handle, operate the handle, and coordinate arm interaction with whole-body motion to open the door against this load and traverse the doorway. As the door rotates, its configuration and the required
interaction direction change during task execution.

A task-level finite-state sequence is used to provide the low-level controller with the corresponding end-effector, force, and locomotion commands. The robot first approaches and grasps the handle in position-control mode. The controller then switches to force control to press the handle and initiate door opening. As the door rotates, the commanded interaction force is gradually redirected from the initial pushing direction toward the tangential direction required to continue opening the door, while the base motion is coordinated with the manipulation behavior. The proposed sensorless force estimate is used as the force-feedback observation of the learned whole-body policy and is also used to identify relevant contact events during task execution. Representative snapshots are shown in Fig.~\ref{fig:hardware_experiment_sequences}(C).

The robot successfully operates the handle and opens the
spring-loaded door while maintaining interaction and coordinating
base and arm motion. This result demonstrates sensorless
force-controlled loco-manipulation against a restoring load
while both the robot and the articulated environment change
configuration. The door-opening experiment therefore demonstrates that the proposed sensorless force-control interface can be incorporated into a higher-level task sequence to execute more complex contact-rich loco-manipulation behaviors. The task-level sequence itself is not the focus of this work; rather, it provides changing interaction commands for evaluating the proposed force-aware whole-body controller.

\subsection{Zero-Force Human-Guidance Experiment}
\label{subsec:human_guidance}

The fourth experiment evaluates the compliant interaction capability of the proposed force-aware controller through human-guided whole-body motion. The desired end-effector interaction force is set to zero, while a human operator applies external forces to the end effector to guide the robot. Representative snapshots are shown in Fig.~\ref{fig:hardware_experiment_sequences}(D).

During the interaction, the proposed sensorless force estimate provides an online measure of the externally applied guidance force. The estimated planar interaction force is converted into a corresponding base-motion command, while the learned whole-body policy coordinates the resulting base and arm motion under a zero desired interaction force. No predefined end-effector trajectory corresponding to the human motion is required.

This experiment complements the active force-control tasks by demonstrating that the same force-aware whole-body control system can also support compliant physical interaction. Together with the valve-rotation, wiping, and door-opening experiments, the results show that explicit sensorless force feedback enables the proposed control system to realize different whole-body physical-interaction objectives.

\section{Conclusions}
\label{sec:conclusion}

This paper presented an RL-based method for force-controlled wheeled-legged loco-manipulation, in which a learned whole-body policy coordinates locomotion, arm motion, and end-effector force regulation under floating-base dynamics. The central idea is to integrate structured interaction-force observation into the policy to support force-guided whole-body contact behavior. To provide this feedback without using an end-effector force/torque sensor for control, we developed a hybrid sensorless estimator that combines generalized momentum observation, contact-constrained wrench projection, and temporal residual compensation.

The estimated force was integrated into a mode-conditioned whole-body RL controller with an axis-wise force/position selector.
This controller can perform free-space end-effector tracking, pure force regulation, and hybrid force/position loco-manipulation within one policy. Simulation results show that the proposed design improves both sensorless force estimation and contact-rich control behavior, enabling accurate force regulation and more effective manipulation of the box toward the desired task state. Hardware valve-rotation and hybrid wiping experiments further demonstrated that the learned controller can use the sensorless force estimate for quantitative force regulation and hybrid force/position interaction. Force-guided door opening and zero-force human-guided motion further demonstrated that the same explicit force-feedback interface can support different forms of contact-rich whole-body interaction. The same policy also operated on hardware with force commands beyond the $\pm10~\mathrm{N}$ training range, demonstrating empirical generalization to larger interaction-force commands and highlighting the practical benefit of providing the estimated interaction force as an explicit observation for learned force regulation.

Despite these results, control performance remains dependent on the fidelity of the robot model, the accuracy of state and contact estimates, and the coverage of the simulation data used to train the residual estimator. Future work will further extend the demonstrated interaction capabilities by integrating the force-aware controller with a high-level planner. This planner will use estimated interaction forces, contact states, and task progress to select manipulation modes, adjust force commands, and sequence more complex loco-manipulation behaviors.

\bibliographystyle{IEEEtran}
\bibliography{References}

\end{document}